\documentclass[11pt,a4paper]{article}
\usepackage{times,latexsym}
\usepackage{url}
\usepackage[T1]{fontenc}

\usepackage[acceptedWithA]{tacl2021v1}

\usepackage{colortbl}
\usepackage[T1]{fontenc}
\usepackage{color,soul}
\usepackage{booktabs}
\usepackage{multirow}
\usepackage{multicol}
\usepackage{diagbox}
\usepackage{booktabs}
\usepackage{amssymb}
\usepackage{xspace}
\usepackage{pifont}
\usepackage{enumitem}

\newcommand{\lt}{\ensuremath <}
\newcommand{\gt}{\ensuremath >}
\newcolumntype{P}[1]{>{\raggedright\arraybackslash}p{#1}}
\newcommand{\specialcell}[2][c]{%
	\begin{tabular}[#1]{@{}c@{}}#2\end{tabular}}

\usepackage{times}
\usepackage{latexsym}
\usepackage{graphicx}
\usepackage{tabularx}
\usepackage{amsmath}
\usepackage{siunitx}
\usepackage{xcolor, soul}
\definecolor{beaublue}{rgb}{0.74, 0.83, 0.9}
\sethlcolor{beaublue}
\definecolor{ForestGreen}{RGB}{34,139,34}

\usepackage{xspace,mfirstuc,tabulary}

\newif\iftaclinstructions
\taclinstructionsfalse 
\iftaclinstructions
\renewcommand{\confidential}{}
\renewcommand{\anonsubtext}{(No author info supplied here, for consistency with
TACL-submission anonymization requirements)}
\newcommand{\instr}
\fi

\iftaclpubformat 

\else

\fi

\title{It's not Rocket Science:\\Interpreting Figurative Language in Narratives}

\author{Tuhin Chakrabarty$^1$\thanks{~~Work done at the Allen Institute for AI}~~~~~Yejin Choi$^{2,3}$~~~~~Vered Shwartz$^{4*}$ \\
 $^1$Columbia University~~~ $^2$Allen Institute for Artificial Intelligence\\
 $^3$Paul G. Allen School of Computer Science \& Engineering, University of Washington\\  
 $^4$University of British Columbia\\
 {\tt\small tuhin.chakr@cs.columbia.edu, yejinc@allenai.org, vshwartz@cs.ubc.ca} \\
}

\begin{document}
\setlength{\parskip}{0pt}
\maketitle
\begin{abstract}
Figurative language is ubiquitous in English. Yet, the vast majority of NLP research focuses on literal language. Existing text representations by design rely on compositionality, while figurative language is often non-compositional. In this paper, we study the interpretation of two non-compositional figurative languages (idioms and similes). We collected datasets of fictional narratives containing a figurative expression along with crowd-sourced plausible and implausible continuations relying on the correct interpretation of the expression. We then trained models to choose or generate the plausible continuation. Our experiments show that models based solely on pre-trained language models perform substantially worse than humans on these tasks. We additionally propose knowledge-enhanced models, adopting human strategies for interpreting figurative language types : inferring meaning from the context and relying on the constituent words' literal meanings. The knowledge-enhanced models improve the performance on both the discriminative and generative tasks, further bridging the gap from human performance.
\vspace{-2ex}

\end{abstract}

\section{Introduction}
\label{sec:intro}
\begin{figure*}[t]
    \centering
    \includegraphics[width=\textwidth]{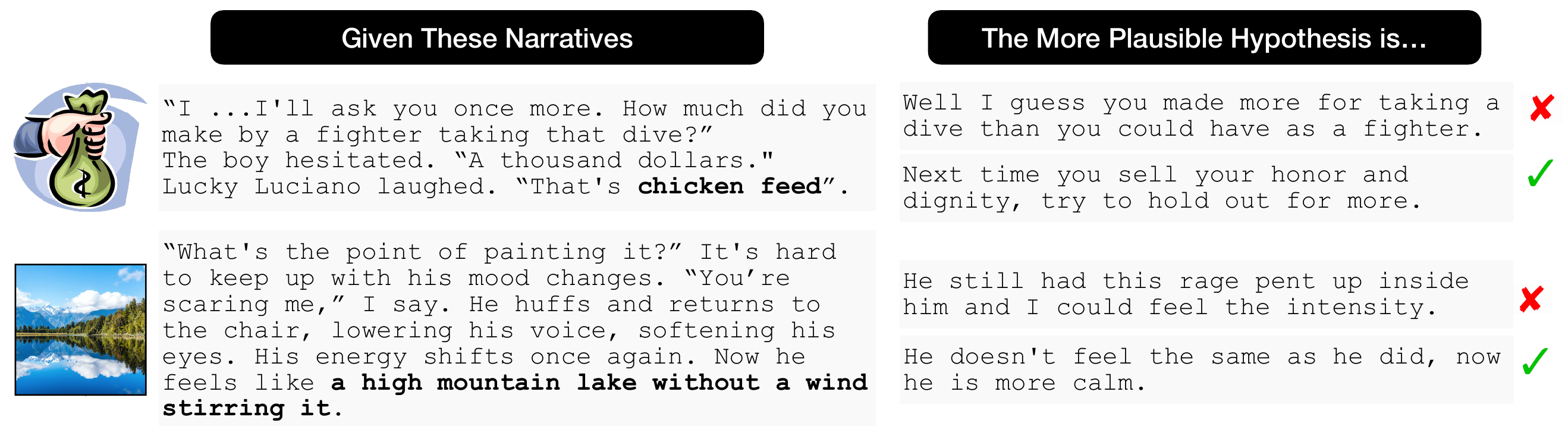}
    \caption{Example narratives from our datasets, containing an idiom (top) or a simile (bottom), along with human-written plausible and implausible continuations.}
    \label{fig:idiomdata}
\end{figure*}

Figurative language is a medium for making language expressive, communicating abstract ideas otherwise difficult to visualize, and provoking emotions \cite{roberts1994people,fussell1998figurative}. Despite the ubiquity of figurative language across various forms of speech and writing, the vast majority of NLP research focuses primarily on literal language. Figurative language is often more challenging due to its implicit nature and is seen as ``a  bottleneck in automatic text understanding'' \cite{shutova2011computational}. 

In recent years, transformer-based language models (LMs) achieved substantial performance gains across various NLP tasks, however, they still struggle with figurative language. In particular, one of the challenges is that figurative expressions are often non-compositional, i.e. the phrase meaning deviates from the literal meanings of its constituents. For instance, the idiom ``chicken feed'' in Figure~\ref{fig:idiomdata} denotes ``a ridiculously small sum of money'' instead of ``food for poultry''. By design, transformer-based LMs compute a word representation as a function of the representation of its context. LM-based phrase representations encode the meanings of the constituent words but hardly capture any meaning that is introduced by the composition itself \cite{yu-ettinger-2020-assessing}. Even though LMs may recognize when a word is used non-literally, and potentially attend to it less, they still struggle to represent the implied, non-literal meaning of such phrases \cite{shwartz-dagan-2019-still}. 

While LMs potentially memorize familiar idioms, we can expect them to further struggle with similes, which are often created ad hoc \cite{carston_wearing_2011}. For example, in Figure~\ref{fig:idiomdata}, the person is compared to ``a high mountain lake without a wind stirring it'' to imply calmness. Many such figurative expressions compose in a non-trivial way, and introduce implicit meaning that requires multiple reasoning steps to interpret.

In this paper we work on interpreting idioms and similes in narratives, where they are especially abundant. Existing work on narrative understanding focuses on literal stories, testing models on their ability to answer questions about a narrative \cite{kocisky-etal-2018-narrativeqa} or continue an incomplete narrative \cite[Story Cloze Test;][]{mostafazadeh-etal-2016-corpus}. We follow the latter setup. We extracted short narratives from the Toronto book corpus \cite{zhu2015aligning}, each containing a figurative expression, and crowdsourced plausible and implausible continuations that rely on correct interpretation of the figurative expression. We defined two tasks: a discriminative setting, where the goal is to choose the plausible continuation among two candidates, and a generative setting, where the goal is to generate a plausible continuation that is coherent with the narrative and complies with the meaning of the figurative expression. 

We report the performance of an extensive number of state-of-the-art LMs on both tasks, in zero-shot, few-shot and supervised settings. Our results show that pre-trained LMs including GPT-3 \cite{brown2020language} perform poorly in the zero-shot and few-shot settings. While the supervised model's performance is closer to humans, the gap is still substantial: in the discriminative tasks, the gap from human performance was 10 and 14.6 points in accuracy for idioms and similes, respectively. In the generative tasks, there was a striking 24 and 28 points difference in human evaluation of the plausibility of generated continuations. 

To further close this gap, we developed knowledge-enhanced models inspired by two human strategies for interpreting unknown idioms, as studied by \newcite{10.2307/3587719} and discussed in \newcite{shwartz-dagan-2019-still}. The first strategy is to infer the expression's meaning from its \emph{context}, for which we incorporate event-centered inferences from ParaCOMET \cite{gabriel-etal-2021-discourse}. The second relies on the \emph{literal} meanings of the constituent words, using concept-centered knowledge from COMET-ConceptNET \cite{Hwang2021COMETATOMIC2O}. Additionally similes are often interpreted by humans using the \emph{literal} property of the vehicle or object of comparison and thus we use concept-centered knowledge here as well. The knowledge-enhanced models consistently outperformed other models on both datasets and settings, with a substantial gap on the generative tasks. 

Furthermore, different strategies were favored for each case: the generative context model performed well on idioms, in line with \citeauthor{10.2307/3587719}'s findings, while the literal model was favored for similes, which are by design based on a constituent's literal attribute (e.g. calm lake). The knowledge-enhanced models leave room for improvement on our dataset. We hope that future work will use additional techniques inspired by the properties of figurative language and human processing of it. Our code and data and code available is available at \url{https://github.com/tuhinjubcse/FigurativeNarrativeBenchmark} 
and our leaderboard is available at \url{https://leaderboard.allenai.org/idiom-simile/}.



\section{Background}
\label{sec:bg}
\subsection{Idioms}
\label{sec:bg_idioms}

Idioms are figurative expressions with a non-literal meaning. For instance, ``break a leg'' is a good luck greeting before a performance and shouldn't be taken literally as wishing someone to injure themselves. Idioms are typically non-compositional, i.e. the meaning of an idiom is not derived from the meanings of its constituents, and fixed, i.e. allowing little variance in syntax and lexical choice.\footnote{The meaning of some idioms may be derived from the non-literal meanings of their constituents. For example, in ``spill the beans'', the non-literal meaning of spill is ``reveal'' and the beans signify the secret \cite{sag2002multiword}.} Idiomatic expressions include proverbs (``actions speak louder than words''), clich\'{e}s (``what goes around comes around''), euphemisms (``rest in peace''), and more. 

Prior work on idioms largely focused on identifying the idiomaticity of a multi-word expression. This is a classification task, defined either at the token-level (is the phrase idiomatic within a given context?), or the type-level (may the phrase be idiomatic in some context?)  \cite{fazly-etal-2009-unsupervised,li-sporleder-2009-classifier,verma-vuppuluri-2015-new,peng-feldman-2016-experiments,salton-etal-2016-idiom,AAAI1714939}. Compared to identification, the interpretation of idioms has been less-explored. Approaches for representing idiomatic expressions include substituting idioms with literal paraphrases \cite{liu-hwa-2016-phrasal,zhou-etal-2021-pie}, representing them as a single token, or learning to compose them at the character rather than word-level \cite{liu-etal-2017-idiom}.

With the rising popularity of pre-trained LMs, several recent papers studied their capacity to accurately represent idioms. \newcite{shwartz-dagan-2019-still} found that while LMs excelled at detecting non-literal word usage (e.g. ``flea'' in ``flea market''), the representation of idiomatic expressions was of lower quality than that of literal ones. \newcite{yu-ettinger-2020-assessing} showed that LMs encode the words that appear in a given text, but capture little information regarding phrase meaning. Finally, \newcite{garcia-etal-2021-assessing} studied the compositionality of noun compounds in English and Portuguese, and found that LM-based models did not perform well on detecting compositionality, and represented idiomaticity differently from humans.

\subsection{Similes}
\label{sec:bg_similes}

Similes are a figure of speech that compares two things, usually with the intent to make the description more emphatic or vivid, and spark the reader's imagination \cite{definition}. Similes may either be explicit, i.e. specify the topic, vehicle, and similarity property, as in ``The house was cold like Antarctica'' (where the topic is ``house", the vehicle is ``Antarctica" and the property of comparison is ``cold"), or implicit, i.e. omitting the property, as in ``the house was like Antarctica'' (Section \ref{sec:datasimile}). Most work in NLP has focused on simile detection, i.e. distinguishing literal from
figurative comparisons. Earlier work relied on semantic and syntactic characteristics, i.e. higher semantic similarity between the topic and the vehicle in literal comparisons than in figurative comparisons \cite{niculae-danescu-niculescu-mizil-2014-brighter,qadir-etal-2015-learning,mpouli-2017-annotating}, and dictionary definitions \cite{qadir-etal-2016-automatically}, while more recent work is based on neural methods \cite{liu-etal-2018-neural,zeng2020neural}. Simile interpretation focused on inferring the implicit property \cite{qadir-etal-2016-automatically}. In other lines of work, \newcite{chakrabarty-etal-2020-generating} and \newcite{Zhang2021WritingPW} proposed methods for generating similes from their literal counterparts, while \newcite{chakrabarty-etal-2021-figurative} showed that state-of-the-art NLI models fail on pragmatic inferences involving similes.  

\subsection{Human Processing of Figurative Language}
\label{sec:processing}

The ways in which humans process figurative language may inspire computational work on figurative language interpretation. \newcite{10.2307/3587719} studied how L2 English speakers interpret unfamiliar English idioms. He found that the leading strategy was to infer the meaning from the given context, which led to successful interpretation in 57\% of the times, followed by relying on the literal meaning of the constituent words (22\% success rate). For example, a participant asked to interpret ``robbing the cradle'' in the context ``Robert knew that he was robbing the cradle by dating a sixteen-year-old girl'' used the literal meaning of cradle to associate the meaning with babies and indirectly with young age, and along with the context inferred that it meant to ``date a very young person''. \newcite{Asl2013TheIO} repeated the same experiment with stories, and concluded that longer contexts improved people's ability to interpret unknown idioms. With respect to novel similes and metaphors, they are interpreted through shared literal attributes between the topic and vehicle (e.g. ``Antarctica is cold, can a house also be cold?'') \cite{Wolff2000EvidenceFR,carston_wearing_2011}. 

\subsection{Narrative Understanding}
\label{sec:bg:narratives}
Early computational work on narrative understanding extracted chains of subevents and their participants from narratives \cite{chambers-jurafsky-2009-unsupervised}. An alternative task is machine reading comprehension, i.e. answering multiple-choice questions based on a narrative, such as MCTest \cite{richardson-etal-2013-mctest} and NarrativeQA \cite{kocisky-etal-2018-narrativeqa}. 

The most commonly used benchmark for narrative understanding today is ROCStories \cite{mostafazadeh-etal-2016-corpus}, a collection of 50k five-sentence commonsense stories pertaining to everyday life. The story cloze task requires models to identify the plausible continuation sentence among two candidate continuations in its discriminative form, or generate a plausible sentence, in its generative form. Since the release of this dataset, many computational approaches for the task were developed \cite[][\emph{inter alia}]{chaturvedi-etal-2017-story,schwartz-etal-2017-story,cai-etal-2017-pay,srinivasan-etal-2018-simple,ijcai2019-249,cui2020discriminative,brown2020language}. In this paper, we follow the story cloze benchmark setup, and collect benchmarks particularly aimed at testing the comprehension of figurative language in narratives. 

\subsection{Commonsense Knowledge Models}
\label{sec:bg:knowledge_models}

Many language tasks require relying on implicit commonsense knowledge that is never mentioned explicitly because it is assumed to be known by everyone. To that end, commonsense knowledge bases (KBs) record such facts. Notably, ConceptNet \cite{speer2017conceptnet} is a large-scale concept-centric KB, while ATOMIC \cite{atomic} contains event-centric knowledge about causes, effects, and the mental states of the participants. To overcome the sparsity of KBs, knowledge models such as COMET \cite{bosselut-etal-2019-comet,Hwang2021COMETATOMIC2O} fine-tuned an LM on structured KB triplets. COMET is capable of providing inferences for new events or concepts. ParaCOMET \cite{Gabriel2021ParagraphLevelCT} is an extension of ATOMIC-COMET which works at the paragraph level and generates discourse-aware commonsense knowledge. Recently several works have used such commonsense knowledge models for improved natural language understanding or generation such as \newcite{bhagavatula2019abductive} for abductive reasoning, \newcite{shwartz-etal-2020-unsupervised} for QA, \newcite{guan2019story,ammanabrolu2020automated} for story generation, \newcite{majumder-etal-2020-like} for dialog generation and \citeauthor{chakrabarty-etal-2020-r}~(\citeyear{chakrabarty-etal-2020-r,chakrabarty-etal-2020-generating,chakrabarty-etal-2021-mermaid}) for creative text generation.

In our work we use the knowledge models COMET \cite{Hwang2021COMETATOMIC2O} and ParaCOMET \cite{Gabriel2021ParagraphLevelCT} respectively to provide more information about the literal meaning of constituent words or the narrative context useful to infer the figurative expressions meaning.

\section{Data}
\label{sec:data}
\begin{table}[t]
\setlength{\tabcolsep}{2pt}
\scriptsize
\centering
\begin{tabular}{ll}
\toprule
\textbf{Idioms} & \textbf{Similes} \\ \midrule
any port in a storm & like a psychic whirlpool\\ 
been there, done that & like a moth-eaten curtain\\  
slap on the wrist & like a first date\\ 
no time like the present & like a train barreling of control\\ 
lay a finger on & like a sodden landscape of melting snow\\  
walk the plank & like a Bunsen burner flame\\  
curry favour & like a moldy old basement\\ 
not to be sneezed at & like a street-bought Rolex\\ 
no peace for the wicked & like an endless string of rosary beads\\

\bottomrule
\end{tabular}
\caption{Examples of idioms and similes present in the narratives in our datasets.}
\label{tab:example_expressions}
\end{table}

We build datasets aimed at testing the understanding of figurative language in narratives, focusing on idioms (Section~\ref{sec:dataidioms}) and similes (Section~\ref{sec:datasimile}). We posit that a model which truly understands the meaning of a figurative expression, like humans do, should be able to infer or decide what happens next in the context of a narrative. Thus, we construct a dataset in the form of the story-cloze test.

\subsection{Idioms}
\label{sec:dataidioms}

We compile a list of idioms, automatically find narratives containing these idioms, and then elicit plausible and implausible continuations from crowdsourcing workers, as follows.

\paragraph{Collecting Idioms.} We compile a list of 554 English idioms along with their definitions from online idiom lexicons.\footnote{\href{http://www.theidioms.com/}{www.theidioms.com},~\href{http://idioms.thefreedictionary.com/}{idioms.thefreedictionary.com}} Table~\ref{tab:example_expressions} presents a sample of the collected idioms.

\paragraph{Collecting Narratives.} We use the Toronto book corpus \cite{zhu2015aligning}, a collection of 11,038 indie ebooks extracted from \url{smashwords.com}. We extract sentences from the corpus containing an idiom from our list, and prepend the 4 preceding sentences to create a narrative. We manually discarded paragraphs that did not form a coherent narrative. We extracted 1,455 narratives with an average length of 80 words, spanning 554 distinct idioms.

\paragraph{Collecting Continuations.} We collected plausible and implausible continuations to the narrative. We used Amazon Mechanical Turk to recruit 117 workers. We provided these workers with the narrative along with the idiom definition, and instructed them to write plausible and implausible continuations that are pertinent to the context, depend on the correct interpretation of the idiom, but which don't explicitly give away the meaning of the idiom. We collected continuations from 3 to 4 workers for each narrative. The average plausible continuation contained 12 words, while the implausible continuations contained 11 words.

To ensure the quality of annotations, we required that workers have an acceptance rate of at least 99\% for 10,000 prior HITs, and pass a qualification test. We then manually inspected the annotations to identify workers that performed poorly in the initial batches, disqualified them from further working on the task, and discarded their annotations.

Our automatic approach for collecting narratives doesn't account for expressions that may be used figuratively in some contexts but literally in others. For example, the idiom ``run a mile'', i.e. avoiding something in any way possible, may also be used literally to denote running a distance of one mile. To avoid including literal usages, we instructed the workers to flag such examples, which we discard from the dataset. We further manually verified all the collected data. Overall we removed 12 such narratives. 

The final idiom dataset contains 5,101 $\lt$narrative, continuation$\gt$ tuples, exemplified in the top part of Figure~\ref{fig:idiomdata}. We split the examples to train (3,204), validation (355) and test (1,542) sets. To test models' ability to generalize to unseen idioms, we split the data such that there are no overlaps in idioms between train and test.

\subsection{Similes}
\label{sec:datasimile}

A simile is a figure of speech that usually consists of a topic and a vehicle (typically noun phrases) which are compared along a certain property using comparators such as ``like'' or ``as'' \cite{hanks2013lexical,niculae-danescu-niculescu-mizil-2014-brighter}. The property may be mentioned (\emph{explicit} simile) or hidden and left for the reader to infer (\textit{implicit} simile). We focus on implicit similes, that are less trivial to interpret than their explicit counterparts \cite{qadir-etal-2016-automatically}, and test a model's ability to recover the implicit property.

\paragraph{Collecting Similes.} Since there are no reliable methods for automatically detecting implicit similes, we first identify explicit similes based on syntactic cues, and then deterministically convert them to implicit similes. We look for sentences in the Toronto Book corpus containing one of the syntactic structures ``as ADJ/ADV as'' or ``ADJ/ADV like'' as a heuristic for identifying explicit similes. We additionally add the constraint of the vehicle being a noun phrase to avoid examples like ``I worked as hard as him''. We remove the adjectival property to convert the simile to implicit, as demonstrated below:


\begingroup
\vspace{2ex}
\renewcommand*{\tabcolsep}{2pt}
{\scriptsize
\hspace{-10pt}
\begin{tabular}{|l|}
\hline
\textbf{Explicit}:\\
He feels \textbf{calm}, like a high mountain lake without a wind stirring it. \\ 
He feels as \textbf{calm} as a high mountain lake without a wind stirring it.\\ \hline
\textbf{Implicit}:\\
He feels like a high mountain lake without a wind stirring it. \\ \hline
\end{tabular}
}
\vspace{2ex}
\endgroup

We collected 520 similes along with their associated property. We asked workers to flag any expression that was not a simile, and manually verified all the collected data. Table~\ref{tab:example_expressions} presents a sample of the collected similes. Many of the similes are original, such as ``like a street-bought Rolex'' which implies that the subject is fake or cheap.

\paragraph{Collecting Narratives.} Once we identified the explicit simile and converted it to its implicit form, we similarly prepend the 4 previous sentences to form narratives. The average length of the narrative was 80 words.

\paragraph{Collecting Continuations.} We repeat the same crowdsourcing setup as for idioms, providing the explicit simile property as the definition. Each narrative was annotated by 10 workers. The average length of continuations was identical to the idiom dataset (12 for plausible and 11 for implausible).

The simile dataset contains 4,996 $\lt$narrative, continuation$\gt$ tuples, exemplified in the bottom part of Figure~\ref{fig:idiomdata}. We split the examples to train (3,100), validation (376) and test (1,520) sets with no simile overlaps between the different sets.

\section{Discriminative Task}
\label{sec:disc}
\begin{figure*}[t]
    \centering
    \includegraphics[width=\textwidth]{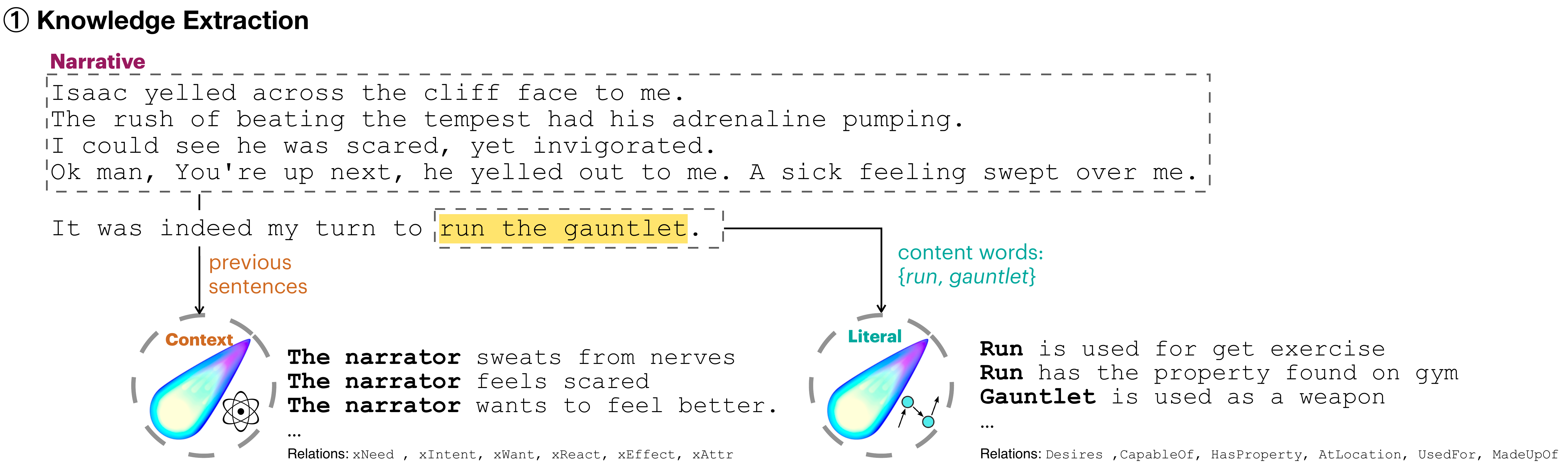} 
    \caption{Extracting inferences from COMET regarding the context (previous sentences in the narrative) and the literal meaning of the content words among the idiom constituents.}
    \label{fig:knowledge_extraction}
\end{figure*}

The first task we derive from our dataset is of discriminative nature in the setup of the story cloze task. Given a narrative N and two candidate continuations $\{\text{C}_1, \text{C}_2\}$, the goal is to choose which of the continuations is more plausible. 

\subsection{Methods}
\label{sec:disc:methods}

For both idioms and similes, we report the performance of several zero-shot, few-shot and supervised methods as outlined below. Most of our experiments were implemented using the transformers package \cite{wolf-etal-2020-transformers}.

\paragraph{Zero-shot.} The first type of zero-shot models is based on standard language model score as a proxy for plausibility. We use GPT-2 XL \cite{Radford2019LanguageMA} and GPT-3 \cite{brown2020language} to compute the normalized log-likelihood score of each continuation given the narrative, predicting the continuation with the highest probability: $ \operatorname{argmax}_i P_{LM}(\text{C}_i |\text{N})$.

We also use UnifiedQA \cite{khashabi-etal-2020-unifiedqa}, a T5-3B model \cite{raffel2020exploring} trained on 20 QA datasets in diverse formats. We don't fine-tune it on our dataset, but instead use it in a zero-shot manner, with the assumption that the model's familiarity with QA format and with the narrative domain through training on the NarrativeQA dataset \cite{kocisky-etal-2018-narrativeqa} would be useful. To cast our task as a QA problem we format the input such that the question is ``Which is more plausible between the two based on the context?''.

\paragraph{Few-shot.} Language models like GPT-3 have shown impressive performance after being prompted with a small number of labelled examples. A prompting example in which the correct continuation is the first is given in the following format: \texttt{Q: N (1) C$_1$ (2) C$_2$ A: (1)}.

We provided the model with as many prompting examples as possible within the GPT-3 API limit of 2,048 tokens, which is 6 examples. The test examples are provided without the answer and the model is expected to generate \texttt{(1)} or \texttt{(2)}. 

We also use the recently proposed Pattern Exploiting Training model \cite[PET;][]{schick-schutze-2021-just}. PET reformulates the tasks as a cloze question and fine-tunes smaller masked LMs to solve it using a few training examples.\footnote{Specifically, it uses ALBERT XXL V2 \cite{lan2019albert}, which is 784 times smaller than GPT-3.} We use the following input pattern: ``\texttt{N. C$_1$. You are \textunderscore}'' for idioms and ``\texttt{N. C$_1$. That was \textunderscore}'' for similes. PET predicts the masked token and maps it to the label inventory using the verbalizer \{``right'', ``wrong''\} for idioms and \{``expected'', ``unexpected''\} for similes respectively mapping  them to \{\textsc{true}, \textsc{false}\}.\footnote{ We also experimented with the pattern and verbalizer used by \newcite{schick-schutze-2021-just} for MultiRC \cite{khashabi-etal-2018-looking}, with the pattern: ``\texttt{N. Question: Based on the previous passage is C$_1$ a plausible next sentence? \textunderscore}.'' and the verbalizer \{``yes'', ``no''\}, but it performed worse.} We provide each model 100 training examples, train it for 3 epochs, and select the model that yields the best validation accuracy.

\paragraph{Supervised.} We fine-tune RoBERTa-large \cite{liu2019roberta} as a multiple-choice model. For a given instance, we feed each combination of the narrative and a continuation separately to the model in the following format: \texttt{N $\lt\text{s/}\gt\text{C}_i$}. 

We pool the representation of the start token to get a single vector representing each continuation, and feed it into a classifier that predicts the continuation score. The model predicts the continuation with the higher score. We fine-tune the model for 10 epochs with a learning rate of \num{1e-5}  and a batch size of 8, and save the best checkpoint based on validation accuracy.

\begin{figure*}[t]
    \centering
    \includegraphics[width=\textwidth]{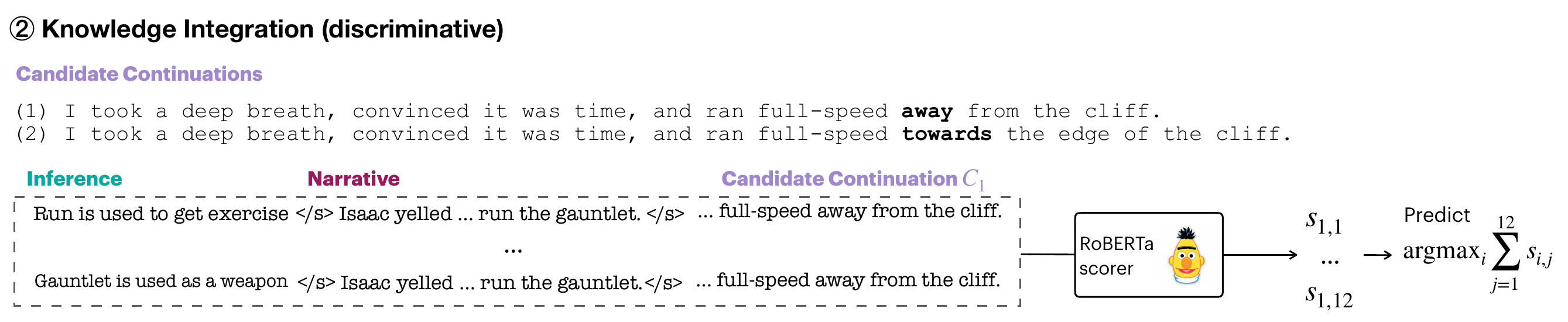}
    \vspace{-4ex}
    \caption{Integrating commonsense inferences into a RoBERTa-based discriminative model.}
    \label{fig:architecture_discriminative}
\end{figure*}

\paragraph{Knowledge-Enhanced.} Inspired by how humans process figurative language, we develop RoBERTa-based models enhanced with commonsense knowledge. We develop two models: the first model obtains additional knowledge to better understand the narrative (\emph{context}), while the second seeks knowledge pertaining to the \emph{literal} meaning of the constituents of the figurative expression (Section~\ref{sec:processing}). In both cases, in addition to the narrative and candidate continuations, the model is also provided with a set of inferences: \{Inf$_1$, ..., Inf$_n$\} that follow from the narrative, as detailed below and demonstrated in Figure~\ref{fig:knowledge_extraction}.

The literal model uses the COMET model \cite{Hwang2021COMETATOMIC2O}, a BART-based language model trained to complete incomplete tuples from ConceptNet. As opposed to extracting knowledge from ConceptNet directly, COMET can generate inferences on demand for any textual input. For an idiom, we retrieve knowledge pertaining to the content words among its constituents, focusing on the following relations: \texttt{UsedFor}, \texttt{Desires}, \texttt{HasProperty}, \texttt{MadeUpOf}, \texttt{AtLocation}, and \texttt{CapableOf}. For each content word, we extract the top 2 inferences for each relation using beam search. For example, given the idiom ``run the gauntlet'', we obtain inferences for ``run'' and ``gauntlet''. We convert the inferences to natural language format based on the templates in \citet{guan2019story}. Given the nature of the simile task, we focused solely on the vehicle's \texttt{HasProperty} relation and obtain the top 12 inferences. For example, given the simile ``like a psychic whirlpool'', we obtain inferences for the phrase ``psychic whirlpool''.

The context model is enhanced with knowledge from ParaCOMET \cite{Gabriel2021ParagraphLevelCT}, trained on ATOMIC. We feed into ParaCOMET all but the last sentence from the narrative, excluding the sentence containing the figurative expression. We generate inferences along ATOMIC dimensions pertaining to the narrator (\texttt{PersonX}), namely: \texttt{xIntent}, \texttt{xNeed}, \texttt{xAttr}, \texttt{xWant}, \texttt{xEffect}, and  \texttt{xReact}. Again, we extract the top 2 inferences for every relation using beam search.

In both models, as demonstrated in Figure~\ref{fig:architecture_discriminative}, the input format $X_{i,j}$ for continuation C$_i$ and inference Inf$_j$ is: \texttt{Inf$_j \lt\text{s/}\gt$ N <s/> C$_i$}.

We compute the score of each of these statements separately, and sum the scores across inferences to get a continuation score:
\begin{equation*}
s_{i} = \sum_{j=1}^{12} s_{i,j} = \sum_{j=1}^{12} \operatorname{scorer}(\operatorname{RoBERTa}(X_{i,j}))
\end{equation*}

\noindent where $\operatorname{scorer}$ is a dropout layer with dropout probability of 0.1 followed by a linear classifier. Finally, the model predicts the continuations with the higher score. We fine-tune the context and literal models for 10 epochs with a learning rate of \num{1e-5} and an effective batch size of 16 for idioms and 64 for similes, and save the best checkpoint based on validation accuracy.

\subsection{Results}
\label{sec:disc:results}

Table~\ref{tab:discriminative_results} shows the performance of all models on the discriminative tasks. For both similes and idioms, supervised models perform substantially better than few-shot and zero-shot models, but still leave a gap of several points of accuracy behind human performance. Human performance is the average accuracy of two native English speakers on the task. We did not provide them with the idiom definition, and we assume they were familiar with the more common idioms. The models performed somewhat better on idioms than on similes, possibly due to the LMs' familiarity with some common idioms as opposed to the novel similes.

Among the zero-shot models, GPT-2 performed worse than GPT-3 and UnifiedQA, each of which performed best on one of the tasks. In particular, UnifiedQA performed well on idioms, likely thanks to its familiarity with the QA format and with the narrative domain.  

In the idiom task, PET outperformed few-shot GPT-3 by a large margin of 12 points in accuracy for idioms and 3.5 points for simile, which we conjecture is attributed to the different number of training examples: 6 for GPT-3 vs. 100 for PET. The small number of examples used to prompt GPT-3 is a result of the API limit on the number of tokens (2,048) as well as the setup in which all prompting examples are concatenated as a single input. 

Overall, few-shot models performed worse than zero-shot models on both datasets. We conjecture that this is due to two advantages of the zero-shot models. First, the GPT-2 and GPT-3 models performed better than the majority baseline thanks to the similarity between the task (determining which continuation is more plausible) and the language model objective (guessing the next word). Second, the UnifiedQA model performed particularly well thanks to its relevant training. At the same time, both few-shot models had to learn a new task from just a few examples.

\begin{table}[t]
\small
\centering
\begin{tabular}{llll}
\toprule
\textbf{Method} & \textbf{Model} & \textbf{Idiom} & \textbf{Simile} \\ \midrule
\multicolumn{2}{l}{Majority} & 50.0 & 50.8 \\ \midrule
\multirow{3}{*}{Zero-shot} & GPT2-XL & 53.6 & 53.7 \\  
 & GPT3 & 60.2 & 62.4 \\ \
 & UnifiedQA & 67.7 & 60.6 \\ \midrule
\multirow{2}{*}{Few-shot} & GPT3 & 54.1 & 51.7 \\ 
 & PET & 66.1 & 55.2 \\ \midrule
\multirow{2}{*}{Supervised} & RoBERTa & 82.0 & 80.4 \\ 
& -narrative & 65.0 & 67.9 \\ \midrule
\multirow{2}{*}{\begin{tabular}[c]{@{}l@{}}Knowledge \\ Enhanced\end{tabular}} & Context & 82.8 & 79.9 \\ 
 & Literal & \textbf{83.5}* & \textbf{80.6} \\ \midrule
\multicolumn{2}{l}{Human Performance} & \textbf{92.0} & \textbf{95.0} \\ \bottomrule
\end{tabular}
\caption{Model performance (accuracy) on the idiom and simile discriminative tasks. $^*$ Difference is significant ($\alpha <0.07$) between the supervised and knowledge-enhanced models via t-test.}
\label{tab:discriminative_results}
\end{table}

The supervised models leave some room for improvement, and the knowledge-enhanced models narrow the gap for idioms. For similes we see a minor drop in the context model and nearly comparable performance for the literal model. 

\paragraph{Annotation Artifacts.} Human-elicited texts often contain stylistic attributes (e.g. sentiment, lexical choice) that make it easy for models to distinguish correct from incorrect answers without solving the actual task \cite{schwartz-etal-2017-effect,cai-etal-2017-pay,gururangan-etal-2018-annotation,poliak-etal-2018-hypothesis}. Following previous work, we trained a continuation-only baseline, which is a RoBERTa-based supervised model that was trained only on the candidate continuations without the narrative. The results in Table~\ref{tab:discriminative_results} (\texttt{-narrative}) show that the performance is above majority baseline, indicating the existence of \emph{some} bias. However, the performance of this baseline is still substantially worse than the supervised baseline that has access to the full input, with a gap of 17 points for idioms and 12 points for similes, indicating that this bias alone is not enough for solving the task.

\subsection{Analysis}
\label{sec:disc:analysis}
 
The knowledge-enhanced models provide various types of inferences corresponding to different relations in ConceptNet and ATOMIC. We are interested in understanding the source of improvements from the knowledge-enhanced models over the supervised baseline, by identifying the relations that were more helpful than others. To that end, we analyze the test examples that were incorrectly predicted by the supervised baseline but correctly predicted by each of the knowledge-enhanced models. We split the examples such that every example consists of a single inference, and feed the following input into the model to predict the plausible continuation: \texttt{Inf <s/> N <s/> C}. We focus on the idiom dataset, since for the literal model for similes the only used relation was \texttt{HasProperty} and the context model performed slightly worse than the baseline. 

\begin{table}[t]
\centering
\small
\begin{tabular}{lrlr}
\toprule
\multicolumn{2}{c}{\textbf{Literal}}  &  \multicolumn{2}{c}{\textbf{Context}} \\ \midrule
\texttt{HasProperty} & 82.3 & \texttt{xNeed} & 82.2 \\ 
\texttt{CapableOf}   & \textbf{83.2} & \texttt{xIntent} & 82.6 \\ 
\texttt{Desires}    & 82.5 & \texttt{xWant} & 82.2 \\ 
\texttt{AtLocation}  & 82.7 & \texttt{xReact} &  \textbf{82.8} \\ 
\texttt{UsedFor}     & 82.4 & \texttt{xEffect} & 82.5 \\ 
\texttt{MadeUpOf}    & 82.8 & \texttt{xAttr} & 82.5 \\ \bottomrule
\end{tabular}
\caption{Percents of successful predictions for each relation type for the test set examples.}
\label{tab:disc_analysis}
\end{table}

Table~\ref{tab:disc_analysis} shows the percents of successful test set predictions for each relation type. The relations in the context model perform similarly, with the best relation \texttt{xReact} performing as well as all of the relations (Table~\ref{tab:discriminative_results}). In the literal model, it seems that the combination of all relations is beneficial, whereas the best relation, \texttt{CapableOf}, performs slightly worse than the full model. For a narrative snippet ``Since Dominic isn't \textit{up for grabs} anymore, I figure that I will concentrate on something else, Carmen declares'', the inference ``grabs is capable of hold on to'' was compliant with the meaning of ``up for grabs'' (available or obtainable), and led to the correct prediction of the plausible continuation ``The good news is that there are many other available bachelors out there''. Conversely, the inference corresponding to the \texttt{Desires} relation was ``grab desires making money'' which was irrelevant and led to an incorrect prediction.

\section{Generative Task}
\label{sec:generative}
\begin{figure*}[t]
    \centering
    \includegraphics[width=1.03\textwidth]{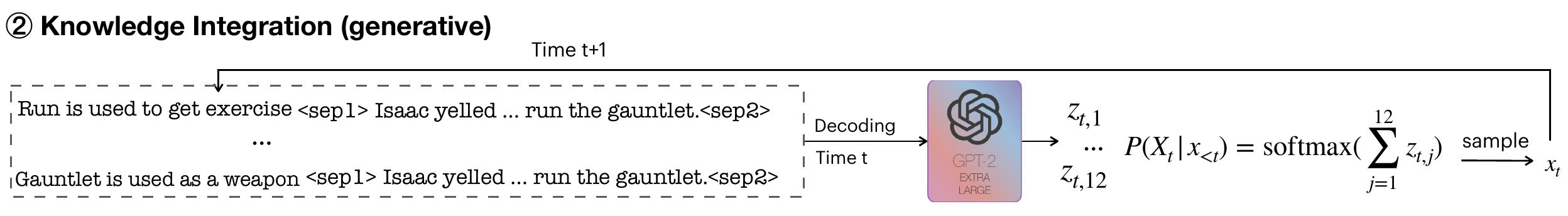}
    \caption{Integrating commonsense inferences into a GPT2-based generative model.}
    \label{fig:architecture_generative}
\end{figure*}

In the generative task, given a narrative N, the goal is to generate a plausible next sentence that is coherent with the context and consistent with the meaning of the figurative expression. Each instance consists of a reference plausible continuation C.

\subsection{Methods}
\label{sec:generative_methods}

We similarly experiment with zero-shot, few-shot, and supervised models. 

\paragraph{Zero-shot.} We use standard LMs, GPT-2 XL and GPT-3, to generate the next sentence following the narrative. We let the models generate up to 20 tokens, stopping when an end of sentence token was generated. Following preliminary experiments, for GPT-2 XL and the rest of the models we use top-k sampling \cite{fan-etal-2018-hierarchical} as the decoding strategy with $k = 5$ and a softmax temperature of 0.7, while for GPT-3 we use the method provided in the API which is nucleus sampling \cite{holtzman2020curious} with a cumulative probability of $p = 0.9$.

\paragraph{Few-shot.} We prompt GPT-3 with 4 training examples of the form \texttt{Q: N A: C} followed by each individual test example, and decode the answer. 

\paragraph{Supervised.} We fine-tune GPT-2 XL with a language model objective for 3 epochs with a batch size of 2. We also trained T5 large \cite{raffel2020exploring} and BART large \cite{lewis-etal-2020-bart} as encoder-decoder models. Both were trained for 5 epochs for idioms and 20 epochs for similes, with an effective batch size of 64. For each model, we kept the best checkpoint based on the validation set perplexity, and used top-k decoding with $k = 5$ and a temperature of 0.7.

\paragraph{Knowledge-Enhanced.} We followed the same intuition and inferences we used for the knowledge-enhanced discriminative models (Section~\ref{sec:disc:methods}). We fine-tune the models for one epoch as the effective data size is multiplied by the number of inferences per sample. The overall architecture of the generative knowledge-enhanced model is depicted in Figure~\ref{fig:architecture_generative}. The models are based on GPT-2 XL and trained with a language model objective to predict the next sentence given the narrative and \emph{a single inference}. The input format for inference Inf$_j$ is: \texttt{Inf$_j$ <sep1> N <sep2>}, where \texttt{<sep1>} and \texttt{<sep2>} are special tokens, and the expected output is the plausible continuation C. During inference, we combine the generations from all inferences pertaining to a given narrative. Inspired by \newcite{liu-etal-2021-dexperts}, who ensemble logits from multiple LMs, we ensemble the logits predicted for multiple input prompts using the same model. 

A standard decoding process gets at each time step an input prompt text $x_{\lt t}$ of length $t-1$. The prompt is encoded and the model outputs the logits for the next ($t^{th}$) token, denoted by $z_{t} \in {\rm I\!R}^{|V|}$, where V is the vocabulary. To get a discrete next token, $z_{t}$ is normalized and exponentiated to resemble a probability distribution over the vocabulary: $P(X_t|x_{\lt t}) = \operatorname{softmax}(z_t)$, and the next token $x_{t}$ is sampled from $P(X_{t}|x_{<t})$. This token is then appended to the prompt and the process iteratively continues until a predefined length or until an end of sentence token had been generated.

Our decoding process differs in that at time step t, we compute the logits ${z_t}_{j=1}^{12}$ corresponding to the prompts derived from each of the inferences: \texttt{Inf$_j$ <sep1> N <sep2>} for $j = 1 ... 12$. We sum the logits vectors to obtain $z_t = \sum_{j=1}^{12} {z_t}_{j}$, from which we decode the next token as usual.

\begin{table}[t]
\renewcommand{\arraystretch}{1.3}
\small
\centering
\begin{tabular}{P{1.3cm}lllll}
\toprule
\multirow{2}{*}{\textbf{Method}} & \multirow{2}{*}{\textbf{Model}} & \multicolumn{2}{l}{\textbf{Idiom}} & \multicolumn{2}{l}{\textbf{Simile}} \\ \cline{3-6}
 & & R-L & B-S & R-L & B-S \\ \midrule
\multirow{2}{*}{Zero-shot} & GPT2-XL & 6.2 & 40.2 & 17.0 & 47.7 \\  
 & GPT3 & 8.2 & 33.6 & 13.9 & 40.2 \\ \midrule
Few-shot & GPT3 & 12.8 & 51.2 & 23.1 & 56.1 \\ \midrule
\multirow{3}{*}{Supervised} & GPT2-XL & \textbf{15.9} & \textbf{54.2} & 26.2 & 59.0 \\ 
 & T5-large & 12.9 & 51.0 & 22.9 & 54.9 \\ 
 & BART-large & 12.4 & 48.8 & 26.7 & 58.4\\ \midrule
\multirow{2}{*}{\begin{tabular}[c]{@{}l@{}}Knowledge\\ Enhanced\end{tabular}} & Context & 15.4 & 52.6 & 20.5 & 55.1 \\  
 & Literal & 13.6 & 51.4 & \textbf{28.9} & \textbf{59.1} \\ \bottomrule
\end{tabular}
\caption{Model performance on the generative tasks in terms of automatic metrics. R-L denotes Rouge-L and B-S denotes BERT-Score.}
\label{tab:generative_results}
\end{table}

\subsection{Results}
\label{sec:generative:results}

\begin{figure*}[t]
    \centering
    \includegraphics[width=\textwidth]{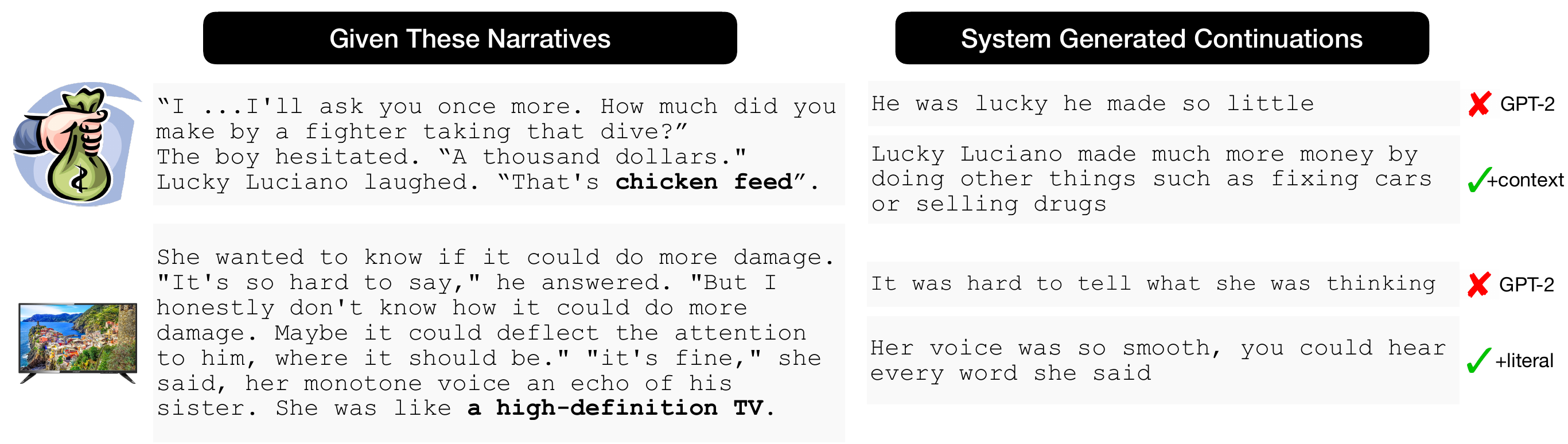}
    \caption{Narratives ending in an idiom (top) or a simile (bottom) with the continuations generated by the baseline GPT-2 model and a knowledge-enhanced model, as preferred by human judges.}
    \label{fig:generations}
\end{figure*}

\begin{table}[t]
\small
\centering
\begin{tabular}{lrrrr}
\toprule
\multicolumn{1}{c}{\textbf{Model}} & \multicolumn{2}{c}{\textbf{\underline{Absolute}}} & \multicolumn{2}{c}{\textbf{\underline{Comparative}}} \\
& \multicolumn{1}{c}{\textbf{Idiom}} & \multicolumn{1}{c}{\textbf{Simile}} &  \multicolumn{1}{c}{\textbf{Idiom}} & \multicolumn{1}{c}{\textbf{Simile}} \\ \midrule
GPT2-XL & 56 & 60 & 15 & 18.6 \\ 
+Context & 68 & 68 & \textbf{45} & 16 \\ 
+Literal & 48 & 76 & 13 & \textbf{46.7} \\ 
Human & \textbf{80} & \textbf{88} & - & - \\ \midrule
All & - & - & 8 & 12 \\ 
Neither & - & - & 17 & 6.7 \\ \bottomrule
\end{tabular}
\caption{Percent of times that the generation from each of the models and human-written references was chosen as plausible (absolute) or preferred (comparative) by the majority of workers.}
\label{tab:humaneval}
\end{table}
\begin{table*}[t]
\centering
\small
\renewcommand{\arraystretch}{1.25}
\begin{tabular}{lcP{12cm}}
\toprule
\textbf{Model} & \textbf{Category} & \textbf{Example} \\ \hline
\multirow{5}{*}{\specialcell{Literal \\(Simile)}}  
 & \multirow{2}{*}{\textcircled{1}} & Is that how you got those bruises on your face? They colored her pure white skin \textbf{like a sloppily designed ink blot test} \textit{It was as if her skin was smudged with a brownie.} \\ \cline{2-3} 
 & \multirow{4}{*}{\textcircled{2}} & 
 We stand there munching next to the evening sun-swept grass and whispering trees of Duboce Park, where a bunch of happy-ass dogs are sprinting around in circles while their owners stand there \textbf{like a scarecrow convention}. \textit{The humans were having a great time, jumping and barking and running around.} \\  \cline{2-3}
\multirow{5}{*}{\specialcell{Context \\(Idiom)}} 
 & \multirow{2}{*}{\textcircled{1}} & She could hold some of her thoughts while she was a wolf.Better than the rest of us. Not as well as Sam, but better than me.She and Derrick were \textbf{thick as thieves} \textit{They would always be looking for opportunities to steal} \\ \cline{2-3} 
 & \multirow{3}{*}{\textcircled{3}} & That's why I packed the heat. Just in case I needed some motivation for him to calm down should he be inclined to \textbf{go ballistic} on me. \textit{{\color{red}: because} I was thinking of ways to solve this problem in a peaceful way} \\ \bottomrule
\end{tabular}
\caption{An example for each error category. Each example consists of a narrative, with the figurative expression in bold and the continuation in italic.}
\label{tab:error_analysis_examples}
\end{table*}

\paragraph{Automatic Evaluation.} Table~\ref{tab:generative_results} shows the performance of all the models on the generative tasks in terms of automatic metrics. We report the performance of the recall-oriented n-gram overlap metric Rouge-L \cite{lin-2004-rouge}, typically used for summarization tasks, and the similarity-based BERT-Score \cite{zhang2019bertscore}. We use the latest  implementation to date which replaces BERT with \texttt{deberta-large-mnli}, which is a DeBERTa model \cite{he2020deberta} fine-tuned on MNLI \cite{williams-etal-2018-broad}. In terms of automatic evaluation, the best-performing knowledge-enhanced model (context for idioms and literal for similes) perform similarly to the GPT-2 XL supervised baseline, with slight preference to the baseline for idioms and to the knowledge-enhanced model for similes. Both types of supervised models outperform the zero-shot and few-shot models.

\begin{table}[t]
\centering
\small
\begin{tabular}{crr}
\toprule
\textbf{Cat.} & \textbf{Literal (Simile)} & \textbf{Context (Idioms)} \\ \hline 
 \textcircled{1} & 50 & 72 \\ 
 \textcircled{2} & 33 & 14 \\
 \textcircled{3} & 17 & 14 \\ \bottomrule
\end{tabular}
\caption{Error categories along with their proportion (in percentage \%) among the implausible continuations.}
\label{tab:error_analysis}
\end{table}

\paragraph{Human Evaluation.}
While automatic metrics provides an estimate of relative model performance, these metrics were often found to have very little correlation with human judgements \cite{novikova-etal-2017-need,krishna2021hurdles}. To account for this we also performed human evaluation of the generated texts for a sample of the test narratives. The human judgements were collected using Amazon Mechanical Turk. Workers were shown a narrative, the meaning of the idiom (or the property of the simile), and a list of 3 generated continuations, one from each of the supervised GPT-2 model, the context model, and the literal model. We performed two types of evaluations. In the absolute evaluation, we randomly sampled 50 narratives for each task, and asked workers to determine for each of the generated continuations along with the human references whether it is plausible or not. In the comparative evaluation, we randomly sampled 100 narratives for idioms and 75 for similes, and presented the workers with a randomly shuffled list of continuations, asking them to choose the most plausible one (or indicate that ``neither of the generations were good'' or ``all are equally good''). In both evaluations, workers were instructed to consider whether the generation is sensical, coherent, follows the narrative, and consistent with the meaning of the figurative expression. Each example was judged by 3 workers and aggregated using majority voting. The inter-annotator agreement was moderate with Krippendorff's $\alpha = 0.68$ and $\alpha = 0.63$ for the absolute and comparative evaluations respectively \cite{Krippendorff2011ComputingKA}. 

In both absolute and comparative performance, Table~\ref{tab:humaneval} shows that for each of the tasks, a knowledge-enhanced model outperformed the baseline GPT-2 model. What makes a more compelling case is that the context model was favored for idioms while the literal model was favored for similes, complying with prior theoretical grounding on these figurative language types. Figure~\ref{fig:generations} shows examples generated by the baseline and the best model for each task. We note that 80\% of the human-written continuations for idioms and 88\% of those in the simile task were judged as plausible. Based on our analysis, the gap from 100\% may be explained by the ambiguity of the narratives that leaves room for subjective interpretation. 

\subsection{Error Analysis}
\label{sec:generative_analysis}

We analyze the continuations labeled as implausible by the annotators, for the best model in each task: context for idioms and literal for similes. We found the following error categories, with percent details in Table~\ref{tab:error_analysis} and exemplified in Table~\ref{tab:error_analysis_examples}: 

\noindent \paragraph{\textcircled{1} Inconsistent with the figurative expression:} The continuation is inconsistent or contradictory to the figurative expression. For instance, the simile in the first row in Table~\ref{tab:error_analysis_examples} is ``like a sloppily designed ink blot test'', for which the property of comparison is ``a pattern of dark blue, purple, and black'', but the generated continuation mentions brownie, which has a \emph{brown} color. Similarly for the idiom ``thick as thieves'' the model generates a literal continuation without understanding its actual meaning ``closest of friends".

\noindent \paragraph{\textcircled{2} Inconsistent with the narrative:} The continuation is inconsistent or contradictory to the flow of the narrative. For instance, the narrative in the second row in Table~\ref{tab:error_analysis_examples} states that ``the owners who are humans are \emph{standing}'', while the continuation states they are jumping. The model further predicts that the \emph{humans} are barking, instead of the \emph{dogs}. In general, across multiple examples we have found that models tend to confuse the various characters in the narrative.

\noindent \paragraph{\textcircled{3} Spelling or grammar errors:} some generations contained spelling mistakes or introduced grammar errors such as starting with a punctuation or having extra blank spaces. Although we instructed the crowdsourcing workers to ignore such errors, they may have affected their plausibility judgements.

\section{Conclusion}
\label{sec:conclusion}
We introduced a narrative understanding benchmark focused on interpreting figurative language, specifically idioms and similes. Following the story cloze test, we designed discriminative and generative tasks with the goal of continuing a narrative. We found that pre-trained LMs irrespective of their size struggle to perform well in zero-shot and few-shot setting, and that the supervised models while competitive are still behind human performance by a significant margin. We further bridged some of this gap with knowledge-enhanced models that are inspired by the way humans interpret figurative expressions. Our analysis reassessed known findings that although LMs generate grammatical human-like texts, they are often inconsistent and the model's ability to distinguish characters in a story is limited. We hope this work will spark additional interest in the research community to further advance the representations and modeling of figurative language, which is too common to ignore. 

\section{Acknowledgement}
This research was supported in part by DARPA under the MCS program through NIWC Pacific (N66001-19-2- 4031), Google Cloud computing, and the Allen Institute for AI (AI2).

\bibliography{anthology,references}
\bibliographystyle{acl_natbib}

\end{document}